# Graphical Estimation of Permeability Using RST&NFIS

H.Owladeghaffari , K.Shahriar
*Department of Mining &Metallurgical Engineering*
*Amirkabir University of Technology*
*Tehran, Iran*
h.o.ghaffari@gmail.com ; k.shahriar@aut.ac.ir

W. Pedrycz
*Department of Electrical and Computer Engineering*
*University of Alberta*
*Alberta, Canada*
pedrycz@ece.ualberta.ca

*Abstract* - **This paper pursues some applications of Rough Set Theory (RST) and neural-fuzzy model to analysis of "lugeon data". In the manner, using Self Organizing Map (SOM) as a pre-processing the data are scaled and then the dominant rules by RST, are elicited. Based on these rules variations of permeability in the different levels of Shivashan dam, Iran has been highlighted. Then, via using a combining of SOM and an adaptive Neuro-Fuzzy Inference System (NFIS) another analysis on the data was carried out. Finally, a brief comparison between the obtained results of RST and SOM-NFIS (briefly SONFIS) has been rendered.**

## I. INTRODUCTION

During the dam structures design, one of the most significant issues is the estimation of permeability variations in different levels of the dam site. However, prediction of permeability, using obtained data, from in-situ tests is a big challenge.

Relating to the determination of potential water flow paths within the rock mass, underlying a potential dam structure is especially important and this has an extensive impact on the planning of grouting procedures [1].

Due to association of uncertainty and vagueness with the monitored data set, particularly, resulted from the in-situ tests (such lugeon test), accounting relevant approaches such probability, Fuzzy Set Theory (FST) and Rough Set Theory (RST) to knowledge acquisition, extraction of rules and prediction of unknown cases, more than the past have been distinguished. The RST introduced by Pawlak has often proved to be an excellent mathematical tool for the analysis of a vague description of object [2], [3].

Application of RST in different fields of the applied sciences has been reported [4], but developing of such system (based on approximate analysis) in rock engineering has not been outstanding, relatively. The main reason is developing and application of numerical modelling (1-1 mapping models), from birthday, in this field. From the other hand, embedding of the rock engineering data with the several of the uncertainties and ambiguities, persuade to consider approximate analysis methods. Under this idea and based on Information Granulation Theory (IGT), we (are) developed (developing) a set of algorithms in relation to the traditional modelling in rock mechanics [5]. Fig 1 shows a general procedure, in which the IGT accompanies by a predefined project based rock engineering design. After determination of constraints and the associated rock engineering considerations, the initial granulation of information as well as numerical (data base) or linguistic format is accomplished. Improvement of modelling instruments based upon IGs, whether in independent or affiliated shape with hard computing methods (such fuzzy finite element, fuzzy boundary element, stochastic finite element…) are new challenges in the current discussion. Thus, one can employ such method as a new methodology in designing of rock engineering flowcharts [5].

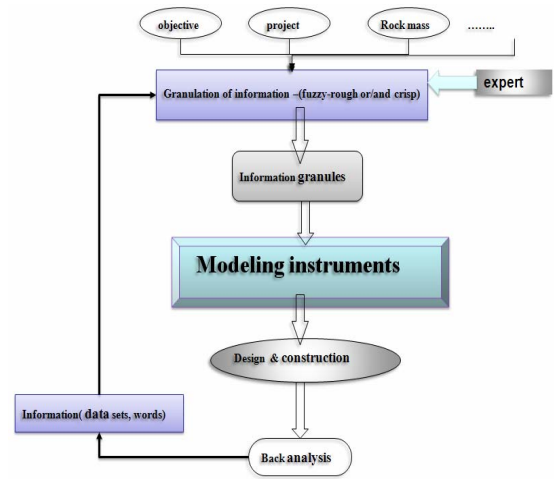

Fig.1 A general methodology for back analysis based on IGT

In this study, according to the "modelling instruments" in fig1, using Self Organizing Map (SOM), NFIS, and RST, some analysis on the permeability data, rounded up from Shivashan dam site located in north western of Iran. In our approach, SOM, NFIS and RST are utilized to construct IGs.

## II. INSTRUMENTS

### A. Self Organizing feature Map (SOM)

Kohonen self-organizing networks (Kohonen feature maps or topology-preserving maps) are competition-based network paradigm for data clustering. The learning procedure of Kohonen feature maps is similar to the competitive learning networks. The main idea behind competitive learning is simple; the winner takes all. The competitive transfer function returns neural outputs of 0 for all neurons except for the winner which receives the highest net input with output 1.

SOM changes all weight vectors of neurons in the near vicinity of the winner neuron towards the input vector. Due to this property SOM, are used to reduce the dimensionality of complex data (data clustering). Competitive layers will automatically learn to classify input vectors, the classes that the competitive layer finds are depend only on the distances between input vectors [6].

*B. Neuro-fuzzy inference system (NFIS)*

There are different solutions of fuzzy inference systems. Two well-known fuzzy modeling methods are the Tsukamoto fuzzy model and Takagi– Sugeno–Kang (TSK) model. In the present work, only the TSK model has been considered.

The TSK fuzzy inference systems can be easily implanted in the form of a so called Neuro-fuzzy network structure .in this study, we have employed an adaptive neuro-fuzzy inference system [7].

*C. Rough Set Theory (RST)*

The rough set theory introduced by Pawlak [2], [3] has often proved to be an excellent mathematical tool for the analysis of a vague description of object. The adjective vague referring to the quality of information means inconsistency, or ambiguity which follows from information granulation.

An information system is a pair S=< U, A >, where U is a nonempty finite set called the universe and A is a nonempty finite set of attributes. An attribute a can be regarded as a function from the domain U to some value set $V_a$. An information system can be represented as an attribute-value table, in which rows are labeled by objects of the universe and columns by attributes. With every subset of attributes $B \subseteq A$, one can easily associate an equivalence relation $I_B$ on U:

$$I_B = \{(x,y) \in U : \text{for every } a \in B, a(x) = a(y)\} \quad (1)$$

Then, $I_B = \bigcap_{a \in B} I_a$.

If $X \subseteq U$, the sets $\{x \in U : [x]_B \subseteq X\}$ and $\{x \in U : [x]_B \cap X \neq \varphi\}$, where $[x]_B$ denotes the equivalence class of the object $x \in U$ relative to $I_B$, are called the *B-lower* and the *B-upper* approximation of X in S and denoted by $\underline{BX}$ and $\overline{BX}$, respectively. Consider $U = \{x_1, x_2, ..., x_n\}$ and $A = \{a_1, a_2, ..., a_n\}$ in the information system S= $\prec$ U, A $\succ$.

By the discernibility matrix M(S) of S is meant an *n\*n* matrix such that

$$c_{ij} = \{a \in A : a(x_i) \neq a(x_j)\} \quad (2)$$

A discernibilty function $f_s$ is a function of *m* Boolean variables $a_1...a_m$ corresponding to attribute $a_1...a_m$, respectively, and defined as follows:

$$f_s(a_1,...,a_m) = \wedge\{\vee(c_{ij}) : i, j \leq n, j \prec i, c_{ij} \neq \varphi\} \quad (3)$$

Where $\vee(c_{ij})$ is the disjunction of all variables with $a \in c_{ij}$. Using such discriminant matrix the appropriate rules are elicited. In this study we have developed dependency rule generation –RST- in MatLab7, and on this added toolbox other appropriate algorithms have been prepared.

III. THE PROPOSED PROCEDURE

Developed algorithms use four basic axioms upon the balancing of the successive granules assumption:
Step (1): dividing the monitored data into groups of training and testing data
Step (2): first granulation (crisp) by SOM or other crisp granulation methods
  Step (2-1): selecting the level of granularity randomly or depend on the obtained error from the NFIS or RST (regular neuron growth)
  Step (2-2): construction of the granules (crisp).
Step (3): second granulation (fuzzy or rough IGs) by NFIS or RST
  Step (3-1): crisp granules as a new data.
  Step (3-2): selecting the level of granularity; (Error level, number of rules, strength threshold...)
  Step (3-3): checking the suitability. (Close-open iteration: referring to the real data and reinspect closed world)
  Step (3-4): construction of fuzzy/rough granules.
Step (4): extraction of knowledge rules

This study involves only SOM and NFIS in the mentioned procedure and other figures of successive granulation can be followed in [8], [9]. Selection of initial crisp granules can be supposed as "Close World Assumption (CWA)" .But in many applications, the assumption of complete information is not feasible, and only cannot be used. In such cases, an "Open World Assumption (OWA)', where information not known by an agent is assumed to be unknown, is often accepted [10]. Balancing assumption is satisfied by the close-open iterations: this process is a guideline to balancing of crisp and sub fuzzy/rough granules by some random/regular selection of initial granules or other optimal structures and increment of supporting rules (fuzzy partitions or increasing of lower /upper approximations ), gradually.

The overall schematic of Self Organizing Neuro-Fuzzy Inference System -Random: SONFIS-R has been shown in fig2. Determination of granulation level is controlled with three main parameters: range of neuron growth, number of rules and error level. The main benefit of this algorithm is to looking for best structure and rules for two known intelligent system, while in independent situations each of them has some appropriate problems such finding of spurious patterns for the large data sets, extra-time training of NFIS for large data set. So, we can use NFIS as an organizing measurement.

It must be noticed by employing a regular neuron growth /rule-or other appropriate parameters- and utilizing other natural computing methods within an intelligent community (network), phase transition of such complex system, in facing with the current debit-I/O-, can be evaluated.

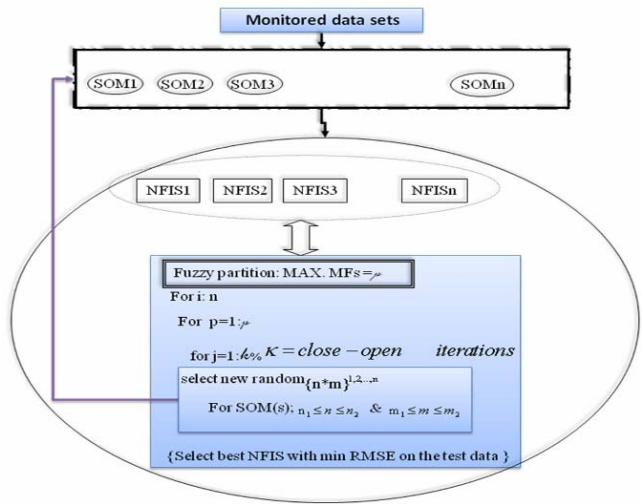

Fig.2 A combining of Self Organizing feature Map and Neuro-Fuzzy Inference System (SONFIS-R)

## IV. RESULTS

Shivashan hydroelectric earth dam is located 45km north of Sardasht city in northwestern of Iran. Geological investigation for the site selection of the Shivashan hydroelectric power plant was made within an area of about 3 square kilometer. The width of the V-shaped valley with similarly sloping flanks, at the elevation of 1185m and 1310m with respect to sea level are 38m and 467m, respectively.

In order to obtain engineering geological information, boreholes were drilled in different points of Shivashan dam's area. Totally, 20 boreholes have been drilled and consequently about 789 objects were resulted. Water Pressure Test (WPT) has used for determination of this area's permeability. WPT is an effective method for widely determination of rock mass permeability. The Lugeon value, which is also known as the Lugeon number (N Lu) is defined as follows: Lu=Water take (liters/meter/min)*[10(bars)/actual test pressure (bars)]

The Lugeon unit is not stated as a ratio of permeability, but to get a sense of proportion, it might be related such that: 1Lugeon=1.3*10-5 cm/s. In practice, usually, the Lugeon test is utilized before grouting to determine quantitatively the volume of water take per unit of time. The maximum meaningful Lugeon is considered 100. A general assessment from all of the boreholes has been shown in fig 3.

To evaluate the permeability due to the lugeon values we follow two situations: 1) utilizing of SONFIS-R and RST-1 on the five chief attributes(fig3); 2) direct application of RST-2 and NFIS on the local coordinates of dam site (as conditional attributes) and lugeon values (as decision part) to depict 3D Iso-surfaces of lugeon variations diagrams.

Analysis of first situation is started off by setting number of close-open iteration and maximum number of rules equal to 10 and 8 (number of rules = 5 to 8) in SONFIS-R, respectively. The error measure criterion in SONFIS is Root Mean Square Error (RMSE), given as below:

$$RMSE = \sqrt{\frac{\sum_{i=1}^{m}(t_i - t_i^*)^2}{m}} \ ;$$

Where $t_i$ is output of SONFIS and $t_i^*$ is real answer; m is the number of test data (test objects). In the rest of paper, let m=93 and number of training data set =600. Figs 4, 5, 7 indicate the results of the aforesaid system. Indicated position in fig 5 states minimum RMSE over the iterations and using 5 rules. Fig 4 shows our mean about structure detection. With 63 neurons in SOM, we acquire some dominant patterns on the problem's space.

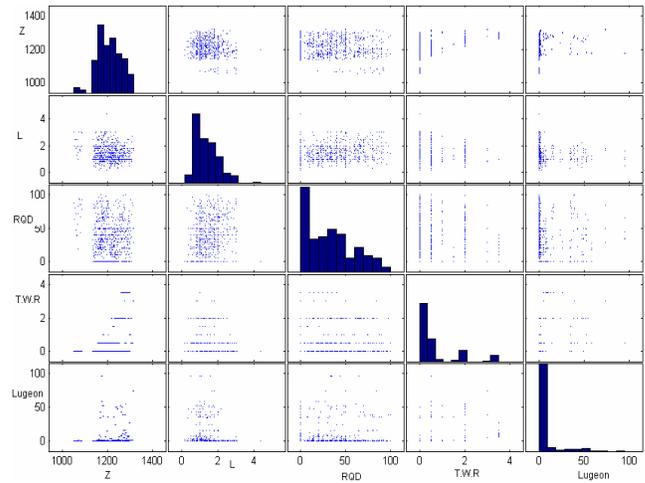

Fig .3 Real data set-Z,L,RQD,T.W.R&lugeon- in matrix plot form (as training data set)

TABLE I
THE REVELED CODES OF TYPE OF WEATHERING ROCK (TWR), MW: MEDIUM WEATHERING, SW: SLIGHTLY WEATHERING, CW: CLAY WEATHERING, HW: HIGH WEATHERING;

| Type of weathering | Ascribed code |
|---|---|
| Fresh-MW | 1.5 |
| SW-MW | 2 |
| Fresh-SW | .5 |
| Fresh | 0 |
| MW | 3 |
| CW | 2.5 |
| SW | 1 |
| HW-MW | 3.5 |
| HW | 4 |

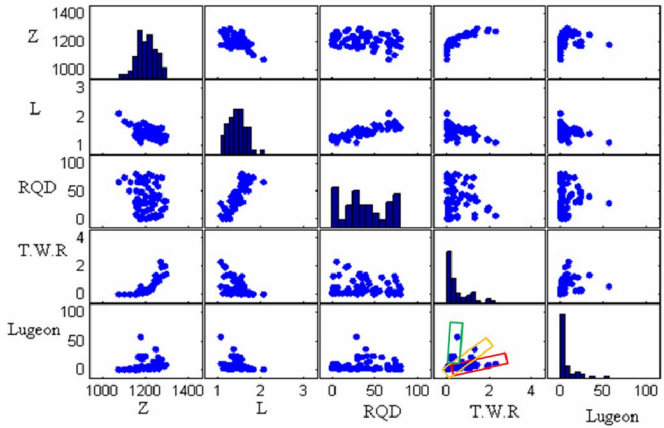

Fig. 4 Matrix plot of crisp granules by 7*9 grid topology SOM after 500 epochs on the training data set (Fig3)

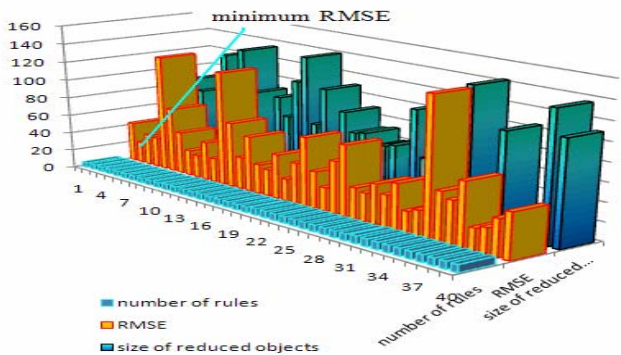

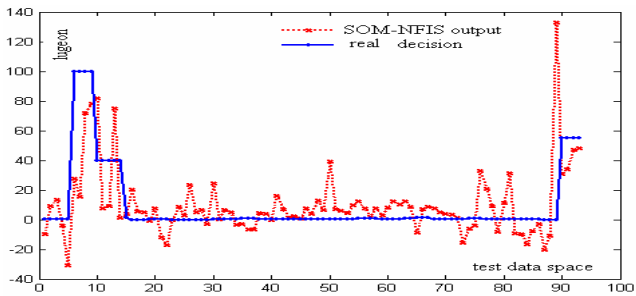

Fig. 5 a) SONFIS-R results with maximum number of rules 8 and close-open iterations 10; b) answer of selected SONFIS-R on the test data

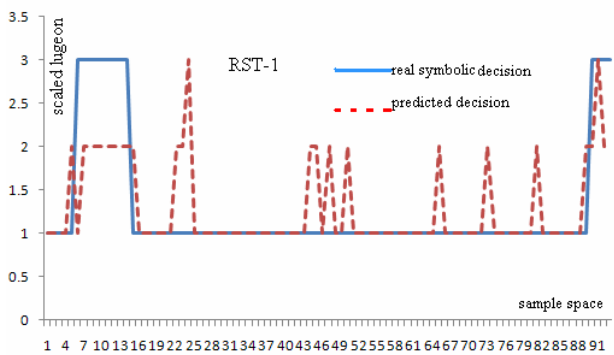

Fig .6 Testing result RST (1) on the test data

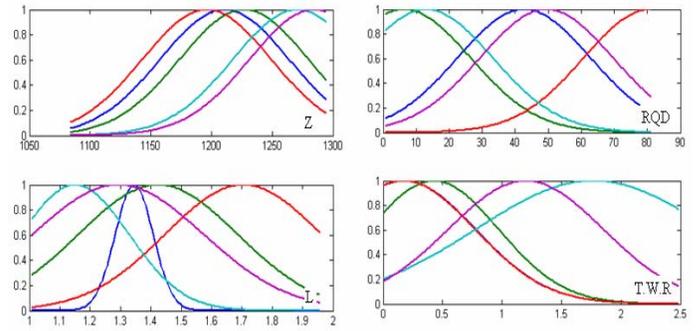

Fig.7 Final membership functions of inputs in SONFIS-R

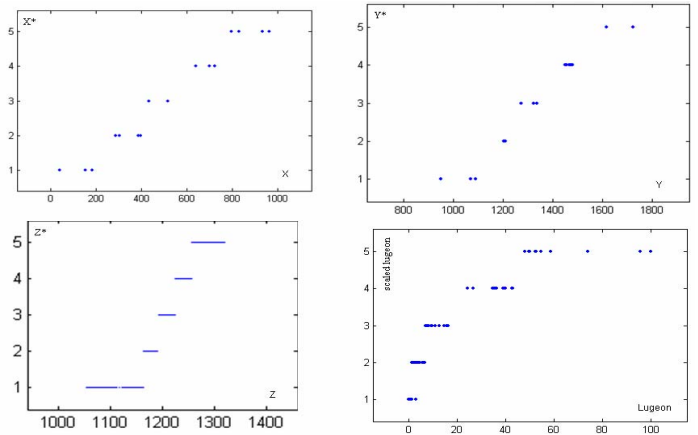

Fig.8 Results of transferring attributes(X, Y, Z and lugeon) in five categories by 1-D SOM

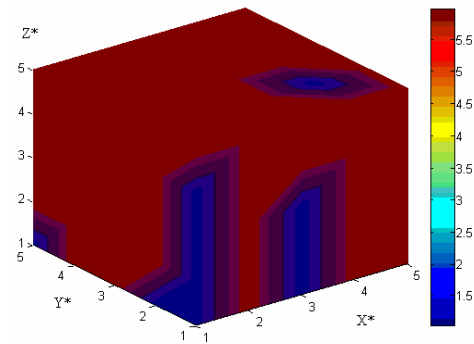

Fig .9 3D views of lugeon variations by RST (2); accomplished by RST (2) and five scaling of attributes. Number 6(more than 5) characterizes ambiguity and unknown cases

With augmenting of close-open iterations SONFIS-R emerges more near min RMSE values, while it is led to the different outcomes, not certainly lowest minimum RMSE. Fig 6 shows the performance of elicited rules by RST-1 on the classification of test data. Total extracted rules on the training data set, in this case, were 63.

Now, we investigate direct application of RST and NFIS on the local coordinates of dam site (as conditional attributes) and lugeon values (as decision part) to depict 3D Iso-surfaces of lugeon variations diagrams. Fig 9 shows the variation of the lugeon data in $Z^*= \{1\}$ to $\{5\}$ which has been acquired by serving five condition attributes in RST (fig 8; the symbolic values by 1-D SOM -5 neurons). The categories 1 to 5 state: very low, low, medium, high, and very high, respectively. Number 6 (more than 5) characterizes ambiguity and unknown cases.

To clarify of permeability changes, in consequent part of rules, the lower value on the symbolic lugeon values which have relatively similar category -for example 1,2,3 or 2,3 or 3,4,5- have been considered. With serving NFIS on such attributes(X, Y, Z& lugeon- without scaling), permeability variations in figs 10, 11 has been portrayed.

In this step, three MFs (Gaussian as like as SONFIS) for input parameters have been utilized. In Consequent of comparison between the results of RST and NFIS, one may interprets the variations, for instance in Z= {2} is the superposition of sub levels, involved Z=1160 to 1200 by NFIS, approximately. So, the compatibility of the results, derived from RST and NFIS can be probed by comparison of fig 9&10. The forecasted domains-gray colours- in fig 9, by RST-2, have been coincided by same regions in fig 10, closely.

It must be noticed that the RST model hasn't covered the high permeability zones, because of employing conservative way in estimation of decision part whereas the NFIS has exposed such possible territories. The rate of lugeon variations, or density of permeable parts, distinguishes the zones with capability of possible spring or hole. Such cavities in the dam structures discussed as "karsts", which are the main characteristics of the limestone deposits (fig 12).

The entire of extracted rules in NFIS is accomplished under subtractive clustering method [7]. To find out the correlation between effective parameters and procuring of valid patterns of the rock mass- in the dam site- one may employ the similar process of NFIS or RST to estimate alterations of RQD and T.W.R (figs 13 and 14 using 3, 5 MFs in NFIS, respectively).

The contrary outputs in some zones with general contextual associated rules about RQD and lugeon, implicate to the relatively complex structures aboard the rock mass. Apart from a few details, comparison of results indicates three overall zones in the rock mass: in first zone the theoretic rules (such reverse relate between RQD& lugeon) are satisfied, but in other zones, the said rule is disregarded. To finding out of the background on these major zones, we refer to the clustered data set by 2D SOM with 7*9 weights in competitive layer (fig 5-b), on the first set of the attributes. The clustered and graphical estimation disclose suitable coordination, relatively. For example in fig 5-b, we have highlighted three distinctive patterns among lugeon and Z, RQD, TWR. One of the main reasons of being such patterns in the investigated rock mass is in the definition of RQD. In measurement of RQD, the direction of joints has not been considered, so that the rock masses with appropriate joints may follow high RQD.

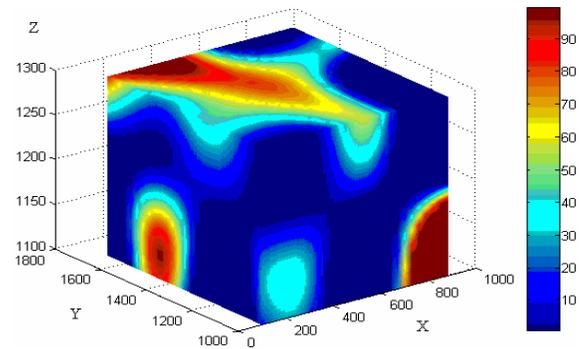

Fig .10 Lugeon variations in z= 1160 to z= 1200; acquired by NFIS

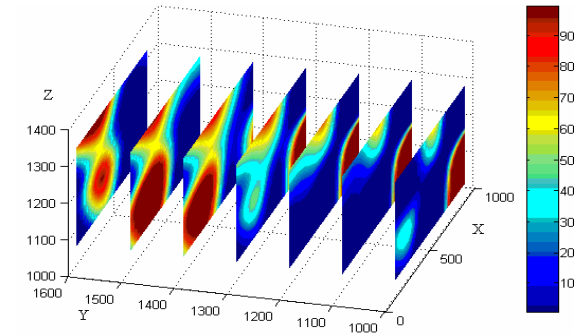

Fig. 11 A cross section perspective of lugeon changes obtained by NFIS

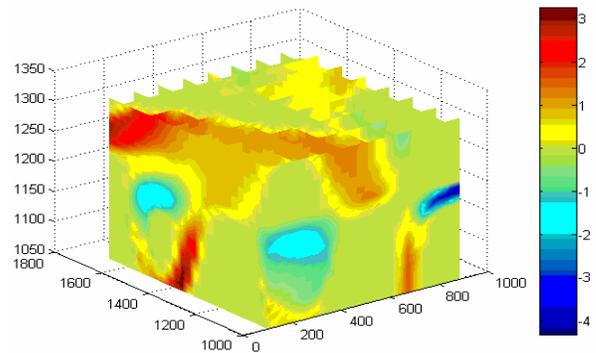

Fig.12 The rate of lugeon variations-possible springs and cavities on the NFIS predictions (divergence of lugeon values)

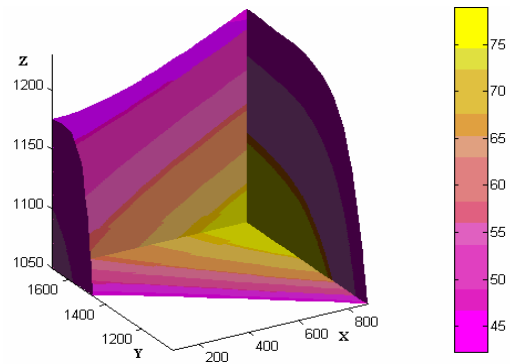

Fig.13 Iso-surfaces of RQD by NFIS

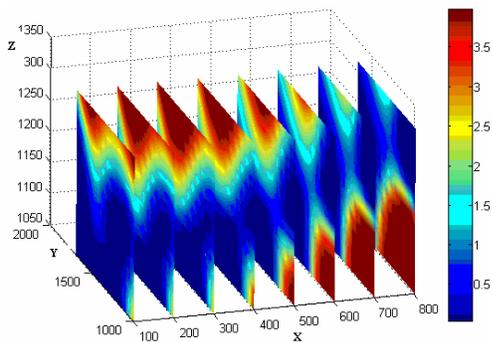

Fig.14 T.W.R variations- X direct cross section view

## V. CONCLUSION

The role of uncertainty in geomechanical information is undeniable feature. Indeed, with developing of new approaches in information theory and computational intelligence, as well as, soft computing approaches, it is necessary to consider these approaches to better understand of natural events in rock mass. Under this view and granulation theory, we proposed two main algorithms, to complete soft granules construction in not 1-1 mapping level of modeling: Self Organizing Neuro-Fuzzy Inference System (Random and Regular neuron growth-, SONFIS-AR- and Self Organizing Rough Set Theory (SORST). So, we used SONFIS-R to analysis of permeability in a dam site, Iran.

So, direct implementation of NFIS and RST on the lugeon data set was proved that the suggested methods could be applied, successfully.  From the mentioned analysis the following results can be deduced:

1- Detection of the permeability variations in successive level using NFIS and RST

2- Elicitation of the dominant simple rules between effective parameters

3- A pre-processing on the scatter lugeon data using best SOM and interpretation of the results